\documentclass[final]{l4dc2026}

\title{Physics-Informed Neural Operators for Cardiac Electrophysiology}
\usepackage{times}
\usepackage{multirow}
\usepackage{multirow}
\usepackage{verbatim}
\usepackage{hyperref}
\usepackage{graphicx}
\graphicspath{ {./Images/} }
\usepackage{xcolor}
\usepackage[normalem]{ulem}
\usepackage{subcaption}

\usepackage{enumitem}
\setlist[itemize]{noitemsep, topsep=2pt, parsep=0pt, partopsep=0pt}
\usepackage[subtle]{savetrees}

\coltauthor{\Name{Hannah Lydon} \Email{hannah.lydon@kcl.ac.uk}\\
 \Name{Milad Kazemi} \Email{milad.kazemi@kcl.ac.uk}\\
\addr Department of Informatics, King's College London, UK \\
\Name{Martin Bishop} \Email{martin.bishop@kcl.ac.uk}\\
\addr Research Department of Digital Twins in Healthcare, King's College London, UK\\
 \Name{Nicola Paoletti} \Email{nicola.paoletti@kcl.ac.uk}\\
\addr Department of Informatics, King's College London, UK
}

\begin{document}
\maketitle
\begin{abstract}%

Accurately simulating systems governed by PDEs, such as voltage fields in cardiac electrophysiology (EP) modelling, remains a significant modelling challenge. Traditional numerical solvers are computationally expensive and sensitive to discretisation, while canonical deep learning methods are data-hungry and
struggle with chaotic dynamics and long-term predictions. 
Physics-Informed Neural Networks (PINNs) mitigate some of these issues by incorporating physical constraints in the learning process, yet they remain limited by mesh resolution and long-term predictive stability. 
In this work, we propose a \textit{Physics-Informed Neural Operator (PINO)} approach to solve PDE problems in cardiac EP.
Unlike PINNs, PINO models learn mappings between function spaces, allowing them to generalise to multiple mesh resolutions and initial conditions. 
Our results show that PINO models can accurately reproduce cardiac EP dynamics over extended time horizons and across multiple propagation scenarios, including zero-shot evaluations on scenarios unseen during training. Additionally, our PINO models maintain high predictive quality in long roll-outs (where predictions are recursively fed back as inputs), and 
can scale their predictive resolution by up to 10$\times$ the training resolution. 
These advantages come with a significant reduction in simulation time compared to numerical PDE solvers, highlighting the potential of PINO-based approaches for efficient and scalable cardiac EP simulations. 
All code used in this work, including experimental results, can be found at \href{https://github.com/janet-9/CardiacEP-PINOS}{https://github.com/janet-9/CardiacEP-PINOS}
\end{abstract}

\section{Introduction}
\label{sec:intro}
Modelling the behaviour of complex dynamical systems governed by partial differential equations (PDEs) is a common problem in engineering, particularly for simulating physical phenomena such as fluid flow and reaction-diffusion systems. Many biological processes can be modelled by PDE systems, and this framing has led to a growth of computational models that aim to simulate the human body to help guide diagnosis and treatment options \citep{katsoulakis_digital_2024}. 

In cardiac healthcare, modelling electrophysiological dynamics poses a key challenge for understanding the origins of potentially dangerous arrhythmias and can aid in planning life-saving treatments, such as catheter ablation or the implantation of cardiac devices. 
To this purpose, several approaches have been explored to simulate cardiac electrophysiology (EP), e.g. \citep{jiang_-silico_2016,trayanova_computational_2024,lydon2025simicd}. 
These methods are often based on mechanistic modelling, such as finite element simulations \citep{plank_opencarp_2021}. As such, they tend to be highly sensitive to mesh resolution and model parameter selection, often resulting in trade-offs between maintaining model accuracy and ensuring computationally efficient simulations. 

Building on the advances and successes of deep learning, learning-based methods have been recently applied to PDE modelling as well, resulting in accurate approximations of the PDE dynamics at a fraction of the simulation/inference cost. 
Physics-informed neural networks (PINNs) \citep{raissi_physics-informed_2019} are a popular technique, which combines a usual data loss term with a physics-based loss to penalise deviations from the (known) laws governing the system. 
In fact, PINNs have already been explored in the field of cardiac EP to model activation sequences in cardiac tissue \citep{sahli_costabal_physics-informed_2020}, as well as more complex cardiac dynamics in both 2D and 3D geometries \citep{herrero_martin_ep-pinns_2022, chiu_physics-informed_2024, nazarov_physics-informed_2022, werneck_replacing_2023, opencarp_pinns_2025}. Whilst PINNs have shown significant promise in replicating the accuracy of traditional finite-element EP models, 
they are limited to simulating on a fixed mesh resolution and often require extensive training budgets to capture the features of the PDE system. Additionally, because PINN models serve as function approximators for a single PDE instance, they require re-training to adapt to unseen model parameters, which limits their generalisation capabilities. 

More recent \textit{Neural Operator (NO)} models 
\citep{kossaifi2024neural, kovachki2021neural} address the above limitations by shifting the training paradigm from fitting a network for one fixed PDE instance (the PINN approach) to learning mappings -- or, operators -- between function spaces, where the input function normally represents an initial condition and the output represents 
the corresponding PDE solution. As such, NO models can learn the dynamics of families of PDEs rather than a single instance (making them independent of the initial condition), can scale efficiently to high dimensions, and are resolution-invariant, i.e., they support inference at different (unseen) discretisation levels. 
Similar to the development of PINNs, NO models have been extended to incorporate physics losses to enforce agreement with known dynamics, an approach called \textit{Physics-Informed Neural Operators (PINOs)} \citep{li_physics-informed_2023}. 

So far, (PI)NO models have primarily been applied to canonical PDE problems such as fluid equations (e.g., Burger's equation and Navier-Stokes model). However, their ability to predict across a family of parametric PDEs (at a fraction of the cost of the original PDE) and generalise across resolutions (akin to numerical methods) makes them an ideal candidate for tackling EP modelling scenarios, which is the main objective of our work. 

\paragraph{Contributions:} This work presents the first physics informed neural operator-based method for the simulation of cardiac electrophysiology scenarios, incorporating known cell models as soft physics constraints with a Fourier neural operator (FNO) \citep{li_fourier_2021} backbone. 
Through extensive experiments, we demonstrate that our approach can:
\begin{itemize}
    \item Produce model predictions of equivalent accuracy to numerical simulations for a range of cardiac wave propagation scenarios.
    \item Perform predictions on unseen data with resolutions significantly higher than those available during training, demonstrating the flexibility of our model.
    \item Achieve high-quality predictions in long roll-out simulations for unseen time horizons. That is, our PINO model also performs well when predictions are made sequentially from past predictions rather than the true inputs, demonstrating its applicability to online prediction scenarios where ground-truth data may be infrequent. 
\end{itemize}

\section{Background}
\label{sec:prelims}

\subsection{Neural Operator Model \label{ssec:neuralops}}

We define our general PDE system in the bounded domain $\mathcal{D}$ as
\begin{equation}\label{eq:pde} 
\frac{\partial u}{\partial t} = \mathcal{R}(u) \text{ on } \mathcal{D} \times (0, \infty),  \quad u = g \text{ on } \partial \mathcal{D} \times (0, \infty),  \quad u = a \text{ on } \overline{\mathcal{D}} \times \{0\}, 
\end{equation}
where we have a partial differential operator $\mathcal{R}$, a known initial condition $a = u(0)$ (where $\overline{\mathcal{D}}$ is the closure of $\mathcal{D}$, all spatial points including the boundary at time $t=0$), and a known boundary condition $g$. For operator learning, we consider a family of PDE models described by Equation \ref{eq:pde} with Banach spaces $\mathcal{A}$ (the space of initial condition functions) and $\mathcal{U}$ (the space of solution functions). For our initial and unknown conditions, we let $a = u(0) \in \mathcal{A}$ and $u(t) \in \mathcal{U}$ for $t > 0$. 

Using this setting, we aim to learn $\mathcal{R}$ using a neural operator model in the method of \citep{kossaifi2024neural, kovachki2021neural}. We take the solution operator to be $\mathcal{G}^{\dagger}: \mathcal{A} \rightarrow \mathcal{U}$ induced by the mapping $a \rightarrow u$. The solution operator $\mathcal{G}^{\dagger}$
can be approximated by the neural operator model represented below:

\begin{equation}
\mathcal{G}_{\theta} = \mathcal{Q} \circ \sigma(\mathcal{W}_{L} + \mathcal{K}_{L} + b_{L}) \circ ... \circ \sigma(\mathcal{W}_{1} + \mathcal{K}_{1} + b_{1}) \circ \mathcal{P},\label{eq:neuralop}
\end{equation}
which consists of two point wise operators $\mathcal{P}: \mathbb{R}^{d_{a}} \rightarrow \mathbb{R}^{d_{L}}$ and $\mathcal{Q}: \mathbb{R}^{d_{L}} \rightarrow \mathbb{R}^{d_{u}}$ such that $\mathcal{P}$ lifts the input function $ a \in \mathcal{A}$ from dimension ${d_{a}}$ to dimension ${d_{L}}$ of the hidden layers, and $\mathcal{Q}$ projects the output of the hidden layers to dimension ${d_{u}}$ of the unknown function $ u \in \mathcal{U}$. The main section of the network consists of $L$ layers of a parametrised pointwise linear operator $\mathcal{W}_{l}$ (a weight matrix), a constant bias $b_l$, and a parametrised kernel operator $\mathcal{K}_{l}$, which are combined and processed by a fixed activation function $\sigma$. All of the parameters of the network are described by $\theta$. The kernel function $\mathcal{K}_{l}$ can take various forms depending on the task. In our case, we consider $\mathcal{K}_{l}$ to be the Fourier convolution operator, resulting in a Fourier Neural Operator model (FNO) as described by \citep{li_fourier_2021}. This kernel operator is described by Equation \ref{eq:fourier} below

\begin{equation}
(\mathcal{K}v^{l})(x) = \mathcal{F}^{-1}(R \cdot (\mathcal{F}v^{l}))(x)
\label{eq:fourier} 
\end{equation}
where $\mathcal{F}, \mathcal{F}^{-1}$ are the fast Fourier transform and its inverse, $v^{l}$ is the function representation under the lifting operator $P$ in layer $l$ of the model, and $R$  represents a matrix of learnable weights. This kernel function can be viewed as a convolution operating in Fourier space, where $R$ assigns different weights to each frequency component. 
The motivation for employing this kernel is that by performing parametrisation in the Fourier domain we can reduce the complexity of the kernel calculations whilst still learning global features, which is ideal for PDE modelling\footnote{To use the fast Fourier transforms, the functions are assumed to be on a uniform and discrete mesh. If not, then an additional transformation (typically a graph kernel operator) can be applied to map the functions to a uniform mesh.}. 

\subsection{Physics-informed Learning}
\label{ssec:physicsloss}
A major challenge in applying machine learning to solve PDE problems is that enforcing physical constraints solely from data is difficult and may require large amounts of data, which can be prohibitive for these systems. The field of physics-informed machine learning addresses this problem by incorporating domain knowledge to guide the model construction. 
Physics-Informed Neural Networks (PINNs) \citep{raissi_physics-informed_2019} are one of the first and most prominent methods. PINNs incorporate physics knowledge as a soft constraint on the neural network by creating loss functions that penalise deviations from the PDE residuals as well as any known boundary and initial conditions.

A general form for this loss function, using $\ell_2$-norm residuals, can be seen in Equation \ref{eq:physics_pinn_loss}, where $u_{\theta}$ denotes the solution function found by the network, $a$ is a given initial condition, $\mathcal{R}$ is the differential operator,  
and $\lambda_{res}$, $\lambda_{bc}$, $\lambda_{ic}$ are weights for each physics loss component\footnote{Both the weightings of the loss components and the parameters of $\mathcal{R}$ can either be fixed or learnt as part of the network parameters, $\theta$.}. 

\begin{equation}
\begin{split}
\mathcal{L}_{\text{phys}}(a, u_{\theta}) = \mathcal{L}_{res} + \mathcal{L}_{IC} + \mathcal{L}_{BC} = \lambda_{\text{res}} \int_0^T \int_D 
\left\| \frac{d u_{\theta}}{dt}(t,x) - \mathcal{R}(u_{\theta}(t,x)) \right\|_2 \, dx \, dt \\
\quad + \lambda_{\text{bc}} \int_0^T \int_{\partial D} 
\left\| u_{\theta}(t,x) - g(t,x) \right\|_2 \, dx \, dt + \lambda_{\text{ic}} \int_D 
\left\| u_{\theta}(0,x) - a(x) \right\|_2 \, dx
\end{split}
\label{eq:physics_pinn_loss}
\end{equation}

As discussed in the Introduction, PINNs can only learn a single PDE instance at a time, requiring retraining with a new instance for each solution function $u\in\mathcal{U}$. Moreover, PINNs rely on discretised meshes to approximate the derivatives required in the PDE residual loss. This requires evaluating the network and its gradients at a large number of collocation points, which scales poorly with both the domain's dimensionality and the complexity of the PDE. Finally, due to their nature as an iterative solver, PINNs can struggle to generalise beyond the spatial and temporal domains seen during training.  

These limitations can be addressed by using operator learning (see Section~\ref{ssec:neuralops}), which, by working over function spaces, can learn a family of PDE instances and generalise across resolutions in a data-efficient manner. 
We next formulate data and physics losses in the NO context. Let $\mathcal{G}_{\theta}$ be the operator model of Equation \ref{eq:neuralop}, learned from a dataset $\{a_{j} , u_{j}=\mathcal{G}^{\dagger}(a_{j})\}^{N}_{j=1}$, where $\mathcal{G}^{\dagger}$ is the true solution operator we seek to approximate. We assume the dataset is induced by a distribution $\mu$ of initial conditions of interest. For a given function pair $(a,u)$, the data loss for $\mathcal{G}_\theta$ is

\begin{equation}\label{eq:no_data} 
\mathcal{L}_\text{data}(a,u) = \int_D \left\| u(x) - \mathcal{G}_\theta(a)(x) \right\|_2 \, dx \
\end{equation}

where the integrals are calculated as the weighted sum of all available spatial points\footnote{In the case that the functions are structured as grids, these weights are equal.}. 

Similarly to PINNs, the final loss is obtained by combining $\mathcal{L}_{\text{data}}$ and $\mathcal{L}_{\text{phys}}$ (see Equation~\ref{eq:physics_pinn_loss}) via weighting parameters $\lambda_{\text{data}}$ and $\lambda_{\text{phys}}$, and averaging over the dataset: 

\begin{equation}
\begin{split}
\mathcal{L} = \mathbb{E}_{a \sim \mu}\left[ \lambda_{\text{data}} \mathcal{L}_\text{data}(a,\mathcal{G}^{\dagger}(a)) +  \lambda_{\text{phys}} \mathcal{L}_{\text{phys}}(a, \mathcal{G}_{\theta}(a)) \right] \approx \\
\frac{1}{N} \left[\sum^{N}_{j=1} \lambda_{\text{data}} \mathcal{L}_\text{data}(a_j,u_j) +  \lambda_{\text{phys}} \mathcal{L}_{\text{phys}}(a_j, \mathcal{G}_{\theta}(a_j)) \right].
\end{split}
\label{eq:no_loss} 
\end{equation}




\section{Problem Setting and Model Design}\label{sec:problem_and_solution}

\subsection{Problem Setting}\label{ssec:problem}
We develop a physics-informed neural operator (PINO) approach for PDE-based cardiac electrophysiology. Previous work have considered this class of systems \citep{herrero_martin_ep-pinns_2022, chiu_physics-informed_2024, opencarp_pinns_2025, nazarov_physics-informed_2022} but using a PINN approach. By leveraging PINOs, we aim to provide a more flexible model able to 1) reproduce different cardiac propagation scenarios without having to retrain the model for each instance and 2) adapt to varying mesh resolutions, thereby drastically reducing both training and runtime costs.

The specific PDE problem we consider concerns simulating 
the propagation of voltage fields across 2D meshes 
using the Aliev-Panfilov cell model \citep{aliev_simple_1996}. This model captures the shape of cardiac action potentials using a reaction diffusion system represented by the following pair of coupled PDEs: 
\begin{align}
\frac{\partial V}{\partial t} = & \ \nabla \cdot (D \nabla V) - kV(V-a)(V-1) - VW
\label{eq:alievpanfilov1} \\
\frac{\partial W}{\partial t} = & \ \left(\epsilon + \frac{\mu_{1} W}{V + \mu_{2}}\right) \left[ -W -kV (V - a -1) \right], 
\label{eq:alievpanfilov2} 
\end{align}
where $V$ represents the transmembrane voltage in scaled dimensionless units, $W$ represents a latent recovery variable and $t$ represents a scaled dimensionless time variable. 
Parameters $k$ and $\epsilon$ represent rate constants for excitation and restitution respectively, $a$\footnote{Not to be confused, in this context, with the operator input $a$ (a voltage map).} represents a threshold parameter for excitation and  $\mu_{1}$ and $\mu_{2}$ are scalable parameters that control the shape of the action potential curve. In particular, $a$ and $\mu_{1}$ can be altered to induce chaotic dynamics in the system, as was demonstrated in \citep{panfilov_spiral_2002}. In our model we consider the domain to be isotropic, which allows the diffusion tensor $D$ to be a scalar value and therefore reduces the first term in Equation \ref{eq:alievpanfilov1} to the Laplacian $D\nabla^{2}V$. Further details on the model implementation, including parameter selection and unit scaling information, can be found in Section~\ref{sec:apmodelextra}.

\subsection{Model Design}
\label{ssec:model}
\paragraph{Model architecture} Our model uses a Fourier Neural Operator (FNO) backbone (see Equations~\ref{eq:neuralop} and~\ref{eq:fourier}) to capture the system's global dynamics. The model architecture, as described in Section~\ref{ssec:neuralops},  consists of an initial grid embedding layer that encodes the positional information of the field maps, a lifting layer to expand to higher dimensions, four FNO blocks consisting of a Fourier convolution kernel, a combination of skip connections and channel MLPs with soft gating modulation, and a projection layer to map the field back to the original domain. 

\paragraph{Data generation} The training data consists of 2D spatio-temporal trajectories of transmembrane voltage $V$ and recovery current $W$ fields (the two variables of our PDE). All samples were processed in batches of dimensionality $[B,D,L,H_{grid},W_{grid}]$, with $B$ being the batch size, $D$ the field dimension\footnote{In our case, the voltage and recovery fields respectively where $d=0$ is the voltage field and $d=1$ is the recovery field.}, $L$ the trajectory length and $H_{grid}\times W_{grid}$ the (spatial) training resolution. 
For data generation, the training time horizon $[0,T]$ -- disjoint from the test horizon -- was discretised using a regular grid, i.e., $0=t_1,\ldots, t_k=T$. We considered two dataset formulations:
\begin{itemize}
    \item \textit{Single Frame Training} ($L=1$): the model is given a single time frame as input and is trained to predict the time frame at $n$ steps ahead. Specifically, for time point $i=1,\ldots,k-n$, we associate a data point $(a,u)$ where $a$ is the PDE solution at $t_i$, i.e., $a=u^{\dagger}(t_i)$, and $u$ is the solution after $n$ steps, i.e., $u=u^{\dagger}(t_{i+n})$.
    \item \textit{Multi Frame training ($L=m$)}: here, the model receives and predicts trajectories of length $m$ instead of individual frames. In this case, for $i=1,\ldots, k-n-2m$, we define the input $a$ as the trajectory $a=(u^{\dagger}(t_i),\ldots,u^{\dagger}(t_{i+m}))$ and its outcome as $u=(u^{\dagger}(t_{i+m+n}),\ldots,u^{\dagger}(t_{i+2m+n}))$.  
\end{itemize}

The reason we consider these two different formulations is to assess whether the predictive quality of the model could be improved by training with small sequences that evolve over time as opposed to single time snapshots, particularly when evaluating the model in sequential roll-outs  (as described in our experiments of Section~\ref{ssec:baselines}). 

\paragraph{Physics loss computation} We note that the choice of frames does not affect the data loss term, as this is defined over the spatial domain only (see Equation~\ref{eq:no_data}), but it does affect the physics-informed loss $\mathcal{L}_\text{phys}$ and specifically, the PDE residual term (Equation~\ref{eq:physics_pinn_loss}), as this requires temporal information to approximate the derivatives. 
In particular, the residual in Equation~\ref{eq:no_data} is approximated using the central difference method, i.e., for time $t_i$ and spatial point $x$:
\begin{equation}\label{eq:phys_loss_finite_diff}
    \frac{d u_{\theta}}{dt}(t_i,x) - \mathcal{R}(u_{\theta}(t_i,x)) \approx \left( \frac{u_{\theta}(t_{i+1})-u_{\theta}(t_{i-1})}{t_{i+1}-t_{i-1}}\right) - \mathcal{R}(u_{\theta}(t_i,x)) 
\end{equation}
where $u_{\theta}(t_{i})$ is the model prediction at time $t_i$, and the term $\mathcal{R}_{\Theta}(u_{\theta}(t_i,x))$ is obtained by plugging the field $u_{\theta}(t_i,x)$ into the RHS of the PDE equations~\ref{eq:alievpanfilov1} and~\ref{eq:alievpanfilov2}. 

The way Equation~\ref{eq:phys_loss_finite_diff} is computed depends on the choice of frame structure. As described in \citep{li_physics-informed_2023}, in the multi-frame approach, the time derivative is calculated within the trajectory, i.e., for sample $b$ in the batch, field dimension $d$, and spatial point $(h,w)$, the difference $u_{\theta}(t_{i+1})-u_{\theta}(t_{i-1})$ is obtained as the prediction indexed by $[b,d,t_{i+1},h,w]$ minus that indexed by $[b,d,t_{i-1},h,w]$. In the single-frame setting, where we have individual points instead of trajectories, the derivative is calculated between consecutive samples of the batch, ensuring first that samples are organised sequentially within the batch: in this case, the above difference is obtained as the prediction indexed by $[t_{i+1},d,1,h,w]$ minus that indexed by $[t_{i-1},d,1,h,w]$.

The residual loss for each of Equations \ref{eq:alievpanfilov1} and \ref{eq:alievpanfilov2} are combined via a weighted sum, with the weights for each residual being scaled in order to bias capturing either the voltage or recovery field.
The boundary conditions in both training methods are modelled as Neumann (no flux) boundary conditions that are enforced on the mesh borders of the output prediction\footnote{This condition, the typical boundary condition used for cardiac electrophysiology modelling, is used to comply with the boundary conditions used in the differential solver that was used to generate the training data, and differs from the general boundary condition in (\ref{eq:pde}). We note that during training, the data is padded at the borders to comply with the assumed periodicity in the FNO.}, 
and the initial condition loss is constructed for the multi-frame approach in order to minimise the errors between the first frame of the ground truth and predicted sample trajectories - that is, minimising the difference between $u^{\dagger}(t_{0})$ and as $u_{\theta}(t_{0})$ to ensure consistency with the ground truth at the start of the predicted trajectory\footnote{This loss term is not included for the single frame training.}

\paragraph{Training algorithm} Training proceeds in two stages: the first stage optimises only the data loss, and is followed by a second stage where both data and physics losses are applied. 
In particular, we explore three solutions to integrate the data and physics loss components during the second stage:
\begin{itemize}
    \setlength{\itemsep}{0pt}
    \setlength{\parskip}{0pt}
    \setlength{\parsep}{0pt}
    \item \textit{Fixed weights}: $\lambda_\text{data}$ and $\lambda_\text{phys}$ remain constant.
    \item \textit{Soft adaptation} \citep{heydari_softadapt_2019}: $\lambda_\text{data}$ and $\lambda_\text{phys}$ are adapted based on their relative magnitudes and rates of change.
    \item \textit{Relative aggregation} \citep{bischof_multi-objective_2022}: $\lambda_\text{data}$ and $\lambda_\text{phys}$ are scaled relative to their previous values, but are occasionally reset to their initial values to avoid bias toward recent fluctuations.
\end{itemize}

\paragraph{Inference} At inference time, we evaluate the prediction quality using 
only the data loss 
(called RMSE in the experimental section). We consider two evaluation settings:
\begin{itemize}
    \item \textit{Point-to-point (P2P)}, where the model receives the ground truth input for prediction. Here, this is equivalent to a single time-step prediction for the single-frame model and a prediction of multiple consecutive time-steps for the multi-frame model.
    \item \textit{Roll-out}, where the model uses past predictions (rather than ground truth) to make new predictions. This setting is obviously more challenging, but serves as a test bed to assess whether our PINO model can be used as a simulator that operates autonomously without true data.
\end{itemize}

\section{Experiments}
\label{sec:Experiments}
\begin{figure}
    \centering
    \scalebox{0.68}{%
        \begin{minipage}[t]{0.23\textwidth}
            \centering
            \includegraphics[width=\textwidth]{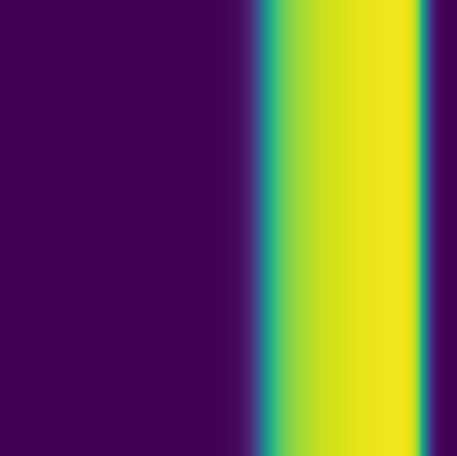}
            \\[-0.3em]
            (a) Planar
        \end{minipage}
        \hfill
        \begin{minipage}[t]{0.23\textwidth}
            \centering
            \includegraphics[width=\textwidth]{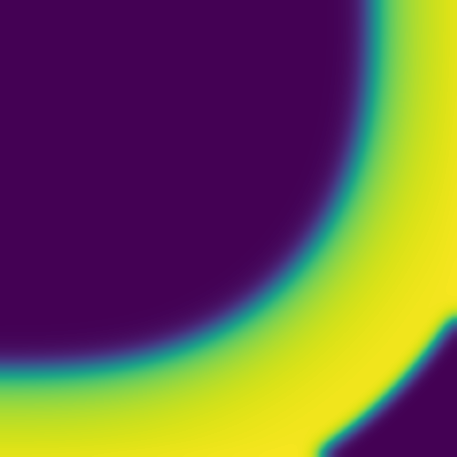}
            \\[-0.3em]
            (b) Centrifugal
        \end{minipage}
        \hfill
        \begin{minipage}[t]{0.23\textwidth}
            \centering
            \includegraphics[width=\textwidth]{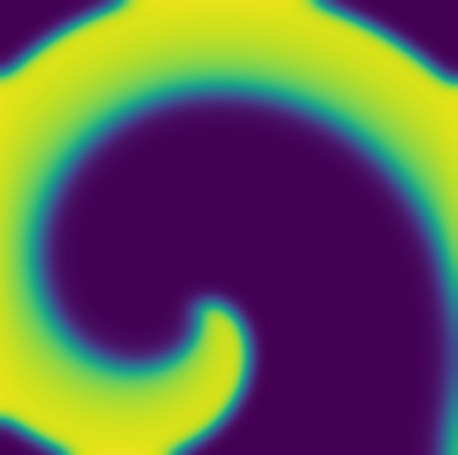}
            \\[-0.3em]
            (c) Spiral
        \end{minipage}
        \hfill
        \begin{minipage}[t]{0.23\textwidth}
            \centering
            \includegraphics[width=\textwidth]{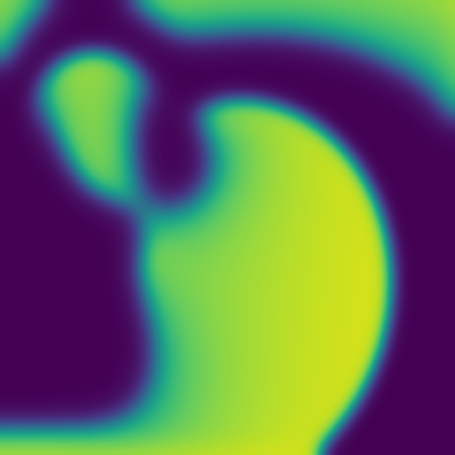}
            \\[-0.3em]
            (d) Break-up
        \end{minipage}
    }
    \caption{Propagation scenarios used for training the operator model, all sampled at $t = 500$ ms.}
    \label{fig:four-scenarios}
\end{figure}

We consider four propagation scenarios for our operator-based simulation: planar, centrifugal (propagation from a single corner of the mesh), spiral and spiral-breakup (pictured in Figure \ref{fig:four-scenarios}). The first two scenarios represent typical propagation patterns in healthy heart scenarios, whilst the latter are characteristic of arrhythmic conditions (with the spiral and spiral-breakup being analogous to monomorphic and polymorphic tachycardia/fibrillation, respectively). Training and evaluation data for all of the propagation scenarios were generated using the open-source cardiac modelling software package openCARP \citep{plank_opencarp_2021}, a high-fidelity PDE simulator for cardiac electrophysiology. Further details of the PDE simulations for data generation can be found in Section~\ref{sec:apmodelextra}. 
 
To create the training sets 
we sampled the trajectories of $1000$ms with a temporal resolution of $dt = 5$ ms and created the input and output pairs for
both single and multi-frame input-output pairs using a moving time window. For the majority of the experiments, we then split the full set of function pairs into a training set representing the first $80\%$ of the trajectory and used the rest of the trajectory as the testing set to encourage the model to extrapolate beyond the training time horizon\footnote{We added a buffer of $2n-1$ frames between the testing and training sets in order to avoid data leakage at the border.}. For the multi-frame training, we used trajectories of length $m=5$ (corresponding to $20$-ms trajectories, or $\sim2.0\%$ of the trajectory).

We trained all of the models using an Adam optimiser with a cosine learning rate scheduler (using an initial learning rate of $5\times10^{-4}$ and a smooth decay over $1000$ epochs). On a GPU-enabled server mounting an NVIDIA A40 48GB, the training time took an average of $\approx$1 hour for single-frame training and $\approx$7 hours for multi-frame training.

\subsection{Baseline Evaluations}
\label{ssec:baselines}
\begin{figure}
  \centering
  \begin{minipage}[b]{\textwidth} %
    \centering
    \includegraphics[width=\textwidth]{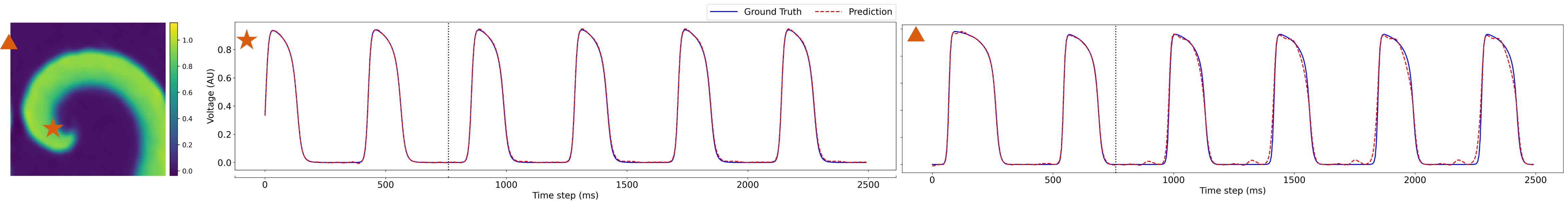}
    \\[-0.3em]
    (a)
  \end{minipage}
  \begin{minipage}[b]{\textwidth} 
    \centering
    \includegraphics[width=\textwidth]{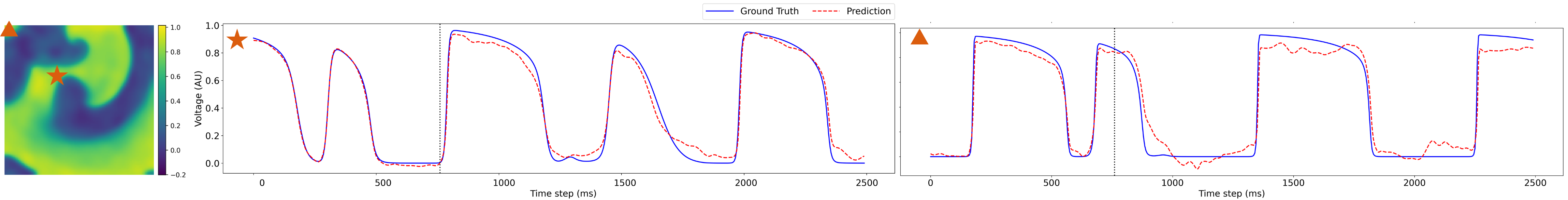}
    \\[-0.3em]
    (b)
  \end{minipage}
  \caption{
    Point-to-point long-time horizon predictions for the (a) spiral and (b) spiral break-up propagation scenarios. The vertical dashed line represents the time horizon for the training data. The best (star) and worst (triangle) performing cells are marked on the voltage maps. The sample snapshots were taken at t=$1700$ms. 
  }
  \label{fig:p2p-stable-chaotic}
\end{figure}

\begin{figure}
  \centering
    \includegraphics[width=\textwidth]{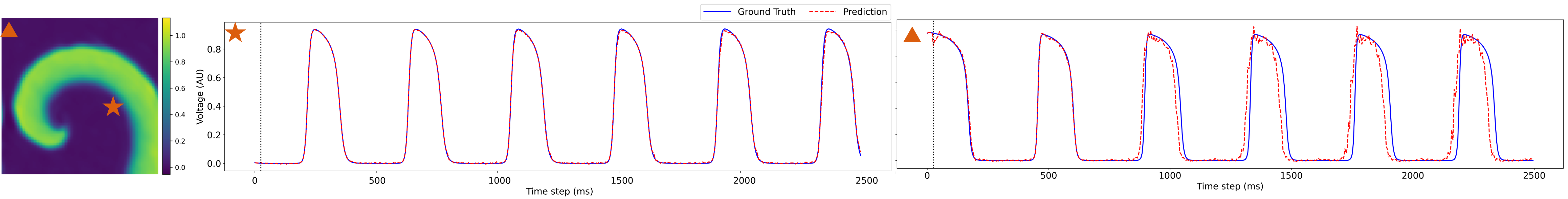}
  \caption{
    Results using the multi-frame roll-out evaluation method for the spiral scenario. The dashed line represents the initial time window given as an input to the model and the best (star) and worst (triangle) performing cells are marked on the sample field, taken at $t=1250$ ms. 
  }
  \label{fig:rollout-stable}
\end{figure}

The first experiment we performed was to evaluate the PINO model described in Section~\ref{sec:problem_and_solution} under baseline conditions, 
(using a fixed mesh resolution for both training and evaluation). The best performing model of the initial training phase, using only data driven losses, was used as the FNO baseline and as the initial state for the secondary training round that incorporates physics losses. The secondary training phases ran for $1000$ epochs (details of variations in training epochs can be found in Section~\ref{ssec:epoch_tests}) 
and, as explained in Section~\ref{sec:problem_and_solution}, we consider three PINO variants based on whether and how the physics loss weights are updated during training\footnote{For the fixed weights PINO model, the losses for the secondary training phase were set as $\lambda_{\text{res}}, \lambda_{\text{ic}}, \lambda_{\text{bc}} = 0.01, 0.1, 0.1$.}. 
Each model was trained in both single-frame and multi-frame modes; and evaluated in a point-to-point (P2P) or roll-out (i.e., sequential) setting.\footnote{We note we evaluated models on the predictions of the transmembrane voltage field only since the recovery field is a latent variable.}
The results are displayed in Table~\ref{tab:baseline_main}, comparing the purely data driven model to the best performing PINO model out of the three training strategies,
and in Figures \ref{fig:p2p-stable-chaotic} and \ref{fig:rollout-stable}. The full table, as well as additional baseline evaluations in which we compare the FNO and PINO model against a PINN, can be found in Section~\ref{ssec:baselines_2}.

\begin{table}
    \centering
    \caption{Baseline comparison for the best performing FNO and PINO models for a fixed mesh resolution.}
    \begin{footnotesize}
    \begin{tabular}{l c c | c || c | c c}
        \hline\hline
        \textbf{Scenario} & \textbf{Model} 
        & \multicolumn{2}{c}{\textbf{Single Frame}} 
        & \multicolumn{2}{c}{\textbf{Multi Frame}} \\
        \cline{3-6}
        & & \textbf{RMSE} & \textbf{RMSE} & \textbf{RMSE} & \textbf{RMSE} \\ 
        & & (P2P) & (Roll-Out) & (P2P) & (Roll-Out) \\
        \hline\hline

        \multirow{4}{*}{Planar} 
            & FNO           & 0.3766 & 0.3819 & 0.3752 & 0.3794 \\
            & PINO          & 0.0095 & 0.2561 & 0.0195 & 0.0204 \\
            \cline{2-6}
            & \textbf{$\Delta$(FNO--PINO)} & \textbf{0.3671} & \textbf{0.1258} & \textbf{0.3557} & \textbf{0.3590} \\
        \hline

        \multirow{4}{*}{Centrifugal} 
            & FNO           & 0.0466 & 0.3826 & 0.0629 & 0.3747 \\
            & PINO          & 0.0178 & 0.3753 & 0.0057 & 0.0112 \\
            \cline{2-6}
            & \textbf{$\Delta$(FNO--PINO)} & \textbf{0.0288} & \textbf{0.0073} & \textbf{0.0572} & \textbf{0.3635} \\
        \hline

        \multirow{4}{*}{Spiral} 
            & FNO           & 0.0248 & 0.4554 & 0.0507 & 0.1682 \\
            & PINO          & 0.0047 & 0.0047 & 0.0113 & 0.0173 \\
            \cline{2-6}
            & \textbf{$\Delta$(FNO--PINO)} & \textbf{0.0201} & \textbf{0.4507} & \textbf{0.0394} & \textbf{0.1509} \\
        \hline

        \multirow{4}{*}{Spiral-Break} 
            & FNO           & 0.0514 & 0.5939 & 0.1144 & 0.3132 \\
            & PINO          & 0.0197 & 0.3961 & 0.0758 & 0.1925 \\
            \cline{2-6}
            & \textbf{$\Delta$(FNO--PINO)} & \textbf{0.0317} & \textbf{0.1978} & \textbf{0.0386} & \textbf{0.1207} \\
        \hline\hline
    \end{tabular}
    \end{footnotesize}
    \label{tab:baseline_main}
\end{table}

The results demonstrate that the PINO approach shows either an equivalent or improved performance compared to the FNO model, with an especially strong improvement for the multi-frame roll-out evaluation, which is the most challenging evaluation scenario. This suggests that the inclusion of physics losses improves the model's ability to make predictions over long time horizons without consistent access to ground truth inputs. 

In addition to the results in Table~\ref{tab:baseline_main} We additionally tested the available models on longer trajectories for the spiral and chaotic datasets, amounting to a full simulation trajectory of $2500$ ms. These results can be found in Section~\ref{ssec:long_horizon} and Figure \ref{fig:p2p-stable-chaotic}.At inference time, the trained model was able to predict a full trajectory via the roll-out method in an average of approximately $1$s for all propagation types.



\subsection{Mesh Resolution Testing}
\label{ssec:meshres}
\begin{figure}
  \centering
  \scalebox{0.8}{ 
    \begin{minipage}{\textwidth}
      \centering
      \begin{minipage}[c]{0.52\textwidth}
        \centering
        \includegraphics[width=\textwidth]{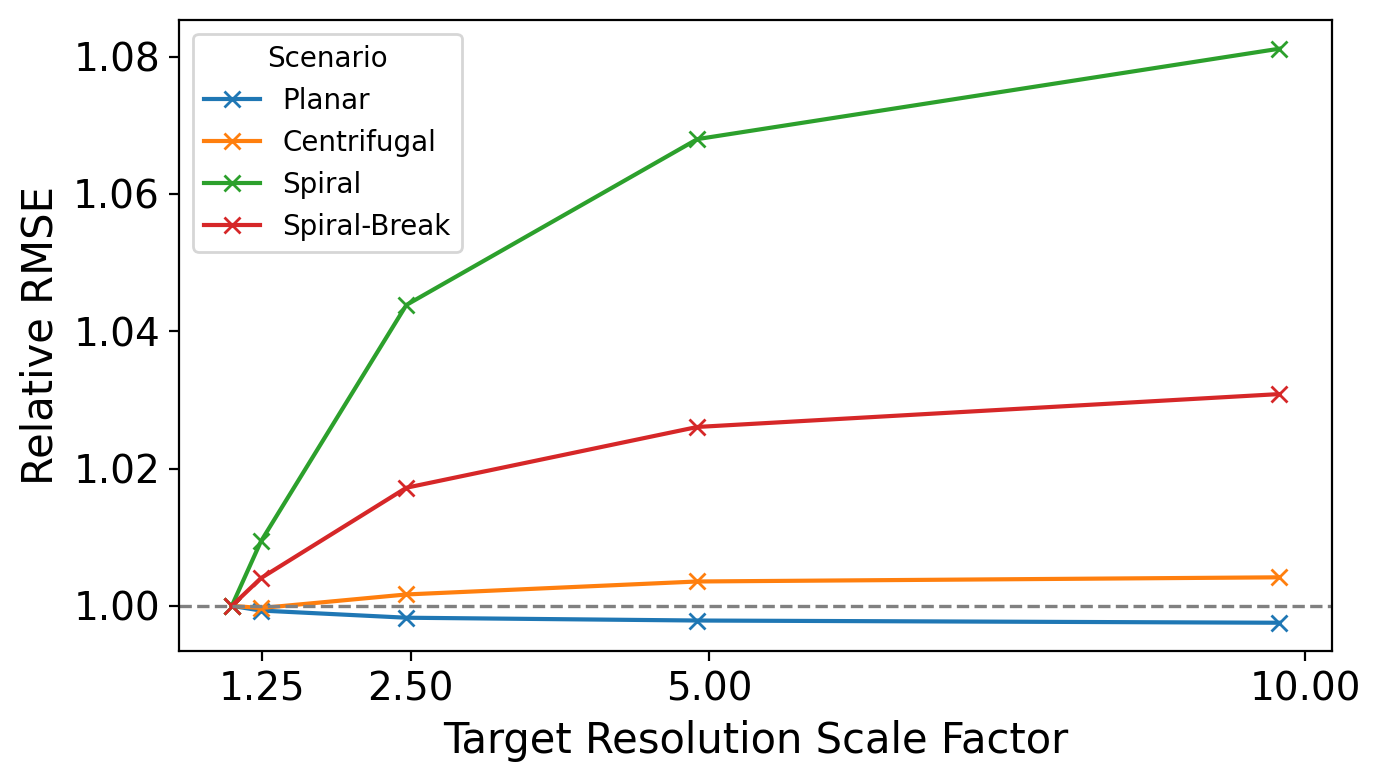}
        \\[-0.3em]
        (a)
      \end{minipage}
      \hfill
      \begin{minipage}[c]{0.2\textwidth}
        \centering
        \includegraphics[width=\textwidth]{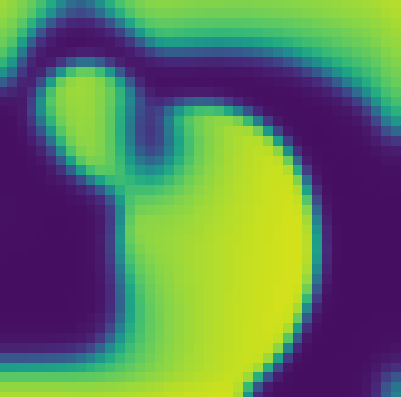}
        \\[-0.3em]
        \raisebox{-0.8em}{(b)} 
      \end{minipage}
      \hfill
      \begin{minipage}[c]{0.21\textwidth}
        \centering
        \includegraphics[width=\textwidth]{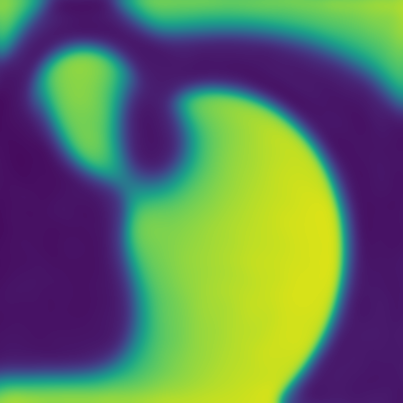}
        \\[-0.3em]
        \raisebox{-0.8em}{(c)} 
      \end{minipage}

    \end{minipage}
  }
  \caption{
    Zero-shot mesh resolution tests.
    (a) Relative RMSE values for P2P evaluation on increasing target resolution scales.
    (b) Input for the chaotic scenario at training resolution ($t = 495$ ms).
    (c) Prediction at the highest evaluation resolution ($t = 500$ ms).
  }
  \label{fig:zero-shot-resolution}
\end{figure}

In order to evaluate the ability of the operator model to extrapolate between different mesh resolutions at inference time, we trained the best-performing PINO model (from the baseline tests) for each scenario on a mesh that was downsampled in spatial resolution by a factor of $10$ from the mesh used to generate the ground truth simulation data. We then evaluated it on target meshes (using P2P evaluation) with resolutions scaled up relative to the training mesh by factors of $1.25$, $2.50$, $5.00$, and $10.00$. As shown in Figure~\ref{fig:zero-shot-resolution}, even when increasing the mesh resolution by a factor of $10$ relative to the training data, the decrease in model accuracy is $< 8\%$ across all propagation scenarios.

\subsection{Zero-Shot Transfer}
\label{ssec:zero_shot}
\begin{figure}
  \centering
  \scalebox{0.9}{
  \begin{minipage}[b]{0.2\textwidth} %
    \centering
    \includegraphics[width=\textwidth]{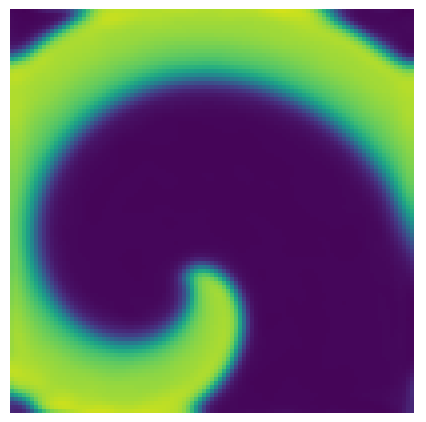}
    \\[-0.3em]
    (a)
  \end{minipage}
  \begin{minipage}[b]{0.2\textwidth} 
    \centering
    \includegraphics[width=\textwidth]{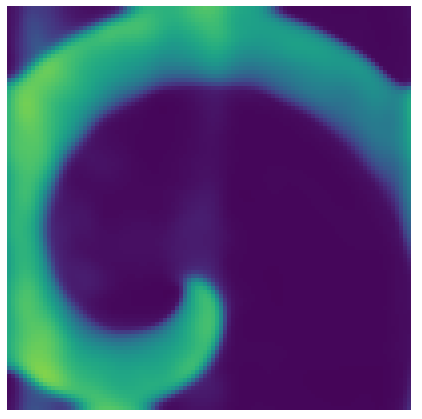}
    \\[-0.3em]
    (b)
     \end{minipage}
    \begin{minipage}[b]{0.235\textwidth} 
    \centering
    \includegraphics[width=\textwidth]{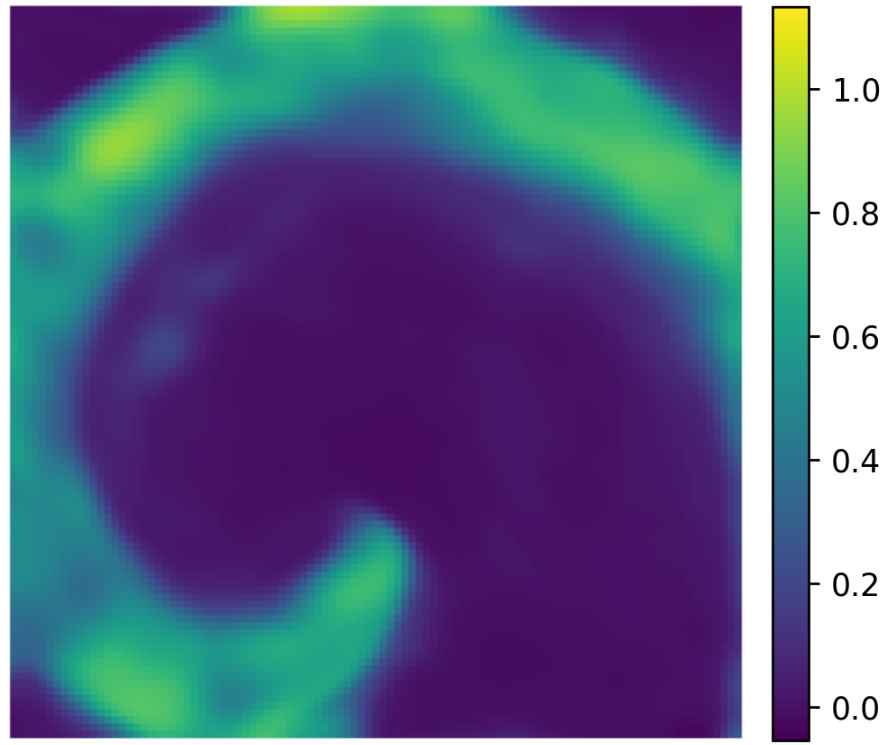}
    \\[-0.3em]
    (c)
     \end{minipage}
     }
  \caption{
    Zero-shot transfer. Samples shown are the predictions for the stable spiral scenario at $t=500$ms using the 
    (a) Stable spiral model.
    (b) Planar model.
    (c) Centrifugal model.
  }
  \label{fig:zero-shot-scenario}
\end{figure}

Finally, we evaluated the performance of our operator model in a zero-shot transfer scenario -- i.e., on datasets that were completely unseen during training -- to test the model's ability to learn the system's underlying dynamics and use that knowledge to predict the trajectory of an unseen propagation scenario. For example, Figure~\ref{fig:zero-shot-scenario} shows the best-performing models trained exclusively on the planar and centrifugal scenarios, evaluated on the (stable) spiral dataset. Despite being trained on simple scenarios, both models can accurately predict the patterns of a more complex propagation scenario. The only difference in the model predictions is in the scaling of the voltage field amplitudes, while the qualitative dynamics remain faithful. Further exploration of the model's generalisation capabilities, demonstrating the ability to learn across varied parametrisation of the cell model and further examples of zero-shot transfer for propagation scenarios can be found in Section~\ref{ssec:generalisation}.  

\section{Conclusion}

In this work, we demonstrated that physics-informed neural operator (PINO) models can efficiently predict cardiac electrophysiology dynamics with comparable accuracy to numerical solvers, with the additional advantage of being able to predict 
across multiple propagation types and parametrisations of the PDE model without retraining. The model was able to perform well in both the single and multiple time step prediction scenarios and also showed promising results when evaluated as a recursive predictor, 
albeit trained using small datasets and modest training budgets. These results demonstrate the strong potential of PINO models for enabling accurate online predictions in resource-constrained medical devices.

Future work will focus on extending this framework to improve the stability of long horizon predictions and to train models using irregular and 3D meshes in order to reflect realistic cardiac anatomy, potentially through employing more recently developed geometry-aware and recurrent operator layers \citep{chen_physics-informed_2025, li_geometry-informed_2023, liu-schiaffini_tipping_2026} into the model to capture more complex spatial patterns and temporal evolution. 
We also aim to 
develop a general operator trained across multiple PDE instances to handle a wider range of cardiac cell models. Although this would increase training costs, strategies such as query-point residuals or temporal-channel separation \citep{diab_temporal_2025} could help maintain computational efficiency.

\bibliography{l4dc2026-refs}
\appendix
\renewcommand{\thesection}{S\arabic{section}} 

\section{Training Data Generation}
\label{sec:apmodelextra}
As described in Section \ref{sec:problem_and_solution}, the Aliev-Panfilov equations model the transmembrane voltage, $V$, the recovery current, $W$, and the time $t$ in atomic units (AU). The scaling for $V$ and $t$ is described in Equation~\ref{eq:alievpanfilovscaling}.
\begin{equation}
V[AU] = \frac{V[mV] + 80}{100}, \hspace{1cm} t[AU] = \frac{t[ms]}{12.9}. 
\label{eq:alievpanfilovscaling} 
\end{equation}

All of the propagation scenarios described in Section~\ref{sec:Experiments} were simulated using the open-source cardiac modelling software package openCARP \citep{plank_opencarp_2021} on a 2D $10\times10 \text{cm}^2$ tetrahedral mesh with a resolution of 250$\mu$m (resulting in a grid of resolution $401\times401$). All of the simulations were performed with isotropic conditions with the monodomain conduction model \citep{brief_history_tissue_models_2014}, and propagation was modelled using the Aliev-Panfilov cell model (using the parameters described in table \ref{tab:ap_params}). 

\begin{table}[h!]
    \centering
    \caption{Parameters used for the Aliev-Panfilov residual loss \\}
    \begin{footnotesize}
    \begin{tabular}{l | c | c | c | c | c | c c}
    \hline
    \hline
    \textbf{Propagation Scenario} & \textbf{$D$} & \textbf{$a$} & \textbf{$\mu_1$} & \textbf{$\mu_2$} & \textbf{$k$} & \textbf{$\epsilon$}\\
    \hline
    \hline
    {Planar, Centrifugal, Spiral} & 0.55 & 0.15  & 0.2 & 0.3 & 8 & 0.002 \\
    \hline
    {Spiral-Break} & 0.55 & 0.099 & 0.1 & 0.3 & 8 & 0.002  \\
    \hline
    \end{tabular}\end{footnotesize}
    \label{tab:ap_params}
 \end{table}
 
To produce the planar and centrifugal waves, the mesh was stimulated from a single wall and a single corner, respectively. Both the spiral and spiral-break-up waves were simulated using a cross-stimulation approach (pacing from a single wall, and then stimulating from a perpendicular wall at a later time in order to induce re-entrance). The simulations were generated with a time horizon of $1000$ms for the planar and centrifugal scenarios in order to capture a single action potential, whilst the spiral and spiral-breakup models were generated with a time horizon of $2500$ms in order to capture the long-term evolution of the system. 

During the construction of the datasets used in training, the voltage field data was normalised to be in AU. When calculating the residual loss, the time was scaled to be in AU. For all propagation scenarios, we used the first $1000$ ms of the simulation to create the training and testing datasets. During simulations, we set $dt = 100$ $\mu$s for the forward solver, which resulted in an average simulation time of $40s$ per $1000ms$ simulation when running the simulation on our GPU server using 12 processing units.  

For the parameter variation experiments described in Section~\ref{sssec:cv_type}, whilst we were not able to scale the parameter of $D$ directly in the simulator, we were able to scale the isotropic conductivity, $g_m$. To relate this scaling to the value of $D$ in the physics loss calculation, we equated the Aliev-Panfilov and monodomain formulations to infer the value of $D$ using the values of membrane capacitance and the cell surface to volume ratio used in the openCARP simulator.

\section{Additional Experiments}
\label{sec:additional_experiments}

\subsection{Model Baseline Evaluations}
\label{ssec:baselines_2}

As referenced in Section~\ref{sec:Experiments}, we conducted a baseline evaluation of the PINO model described in Section~\ref{sec:problem_and_solution} against both a PINN and a purely data-driven FNO, using a fixed mesh resolution for both training and evaluation. For these evaluations, each propagation scenario was considered separately and 
the PINN was trained using the first 80$\%$ of each trajectory and tested on the last 20$\%$ to match the training split for the operator models. The training budgets for all models were on the order of $10^{7}$ parameters; each model was trained for $1000$ epochs with the specific propagation type, and the best-performing model was saved based on the MSE score on the test set. The error metrics
for each model are shown in Table \ref{tab:baseline_model_comparison}, which compares the baseline results for the FNO and PINO for single frame P2P prediction (as also shown in Tables~\ref{tab:baseline_main} and \ref{tab:baseline_full}), to the PINN model.

\begin{table}[h]
\centering
\caption{Performance comparison of RMSE scores across models for each propagation scenario. 
}
\begin{footnotesize}
\begin{tabular}{l | c | c | c | c}
\hline\hline
\textbf{Model}
 & \multicolumn{1}{c|}{\textbf{Planar}} 
 & \multicolumn{1}{c|}{\textbf{Centrifugal}} 
 & \multicolumn{1}{c|}{\textbf{Spiral}} 
 & \multicolumn{1}{c}{\textbf{Breakup}} \\
\hline\hline

PINN 
& 0.1114 
& 0.1436
& 0.4593 
& 0.3011 \\

FNO 
& 0.3766 
& 0.0466 
& 0.0248 
& 0.0514 \\

{PINO}
& 0.0095
& 0.0178 
& 0.0047 
& 0.0197 \\

\hline\hline
\end{tabular}
\end{footnotesize}
\label{tab:baseline_model_comparison}
\end{table}

Compared to the operator models, the PINN's performance was generally worse than the operator models when trained 
with an equivalent training budget. It is likely that the PINN model required more extensive training to fully learn the system dynamics, particularly for the more complex propagation scenarios -- highlighting the operator model's capability to learn system dynamics with a more computationally efficient training budget than neural network-based approaches. 

\begin{table}
    \centering
    \caption{Baseline comparison for the FNO and all PINO models for a fixed mesh resolution, with the best performing model highlighted in bold.
    \\ ($^{*}$ indicates cases in which the model prediction collapsed.)}
    \begin{footnotesize}
    \begin{tabular}{l c c | c || c | c c}
        \hline\hline
        \textbf{Scenario} & \textbf{Model} 
        & \multicolumn{2}{c}{\textbf{Single Frame}} 
        & \multicolumn{2}{c}{\textbf{Multi Frame}} \\
        \cline{3-6}
        & & \textbf{RMSE} & \textbf{RMSE} & \textbf{RMSE} & \textbf{RMSE} \\ 
        & & (P2P) & (Roll-Out) & (P2P) & (Roll-Out) \\
        \hline\hline

        \multirow{5}{*}{Planar} 
            & FNO           & 0.3766 & 0.3819 & 0.3752 & 0.3794 \\
            & PINO-Fixed     & 0.0257 & 0.4735 & 0.0276 & 0.0442 \\
            & PINO-SoftAdapt & 0.0120 & 1.9414$^{*}$ & 0.0211 & \textbf{0.0204}\\
            & PINO-RelAdapt  & \textbf{0.0095} & \textbf{0.2561} & \textbf{0.0195} & 0.3776 \\
        \hline

        \multirow{5}{*}{Centrifugal} 
            & FNO           & 0.0466 & 0.3826 & 0.0629 & 0.3747 \\
            & PINO-Fixed     & 0.0415 & \textbf{0.3753} & 0.0072 & 0.0146 \\
            & PINO-SoftAdapt & 0.0253 & 0.4661 & \textbf{0.0057} & \textbf{0.0112}\\
            & PINO-RelAdapt  & \textbf{0.0178} & 0.4027 & 0.0129 & 0.1418 \\
        \hline

        \multirow{5}{*}{Spiral} 
            & FNO           & 0.0248 & 0.4554 & 0.0507 & 0.1682 \\
            & PINO-Fixed     & 0.0060 & 0.0304 & \textbf{0.0113} & 0.0176 \\
            & PINO-SoftAdapt & 0.0059 & 0.0295 & 0.0117 & 0.0179 \\
            & PINO-RelAdapt  & \textbf{0.0047} & \textbf{0.0047} & \textbf{0.0113} & \textbf{0.0173} \\
        \hline

        \multirow{5}{*}{Spiral-Break} 
            & FNO           & 0.0514 & 0.5939 & 0.1144 & 0.3132 \\
            & PINO-Fixed     & 0.0410 & 0.6004 & 0.0766 & \textbf{0.1925} \\
            & PINO-SoftAdapt & 0.0432 & 0.5427 & \textbf{0.0758} & 0.1993 \\
            & PINO-RelAdapt  & \textbf{0.0197} & \textbf{0.3961} & 0.0832 & 0.3647 \\
        \hline\hline
    \end{tabular}
    \end{footnotesize}
    \label{tab:baseline_full}
\end{table}

\subsection{Long Horizon Predictions}
\label{ssec:long_horizon}
As described in Section~\ref{ssec:baselines}, we evaluated the performance of the spiral and spiral break-up models on long time horizon trajectories, extending the time horizon of the training trajectory by a factor of approximately $3$. The results tend to be in agreement with Table~\ref{tab:baseline_main} in that the PINO model is generally equivalent to or better than the FNO model. 
As expected, given the greater challenge of predicting over a longer time horizon, the RMSE is generally higher than in the baseline results, particularly for the spiral break-up model. This is primarily due to the difficulty of predicting chaotic systems over long time horizons, and suggests that enforcing the PDE as a soft constraint for learning wasn't sufficient to capture the evolved dynamics of the system. Visualisations of the P2P results can be seen in the main body of the text in Figure~\ref{fig:p2p-stable-chaotic}.

\begin{table}[h!]
     \centering
    \caption{Long-Horizon Baseline performances for a fixed mesh resolution \\ ($^{*}$ indicates cases in which the model prediction collapsed) 
    }
    \begin{footnotesize}
    \begin{tabular}{l c c | c || c | c c}
    \hline
    \hline
    \textbf{Scenario} & \textbf{Model} 
    & \multicolumn{2}{c}{\textbf{Single Frame}} 
    & \multicolumn{2}{c}{\textbf{Multi Frame}} \\
    \cline{3-6}
     &  & \textbf{RMSE} & \textbf{RMSE} 
        &  \textbf{RMSE} & \textbf{RMSE} \\
     &  &  (P2P) & (Roll-Out) &  (P2P) & (Roll-Out) \\
    \hline
    \hline

    \multirow{4}{*}{Spiral} 
        & FNO     & 0.0270 &  0.5294 & 0.0525 &  0.2733 \\
        & PINO-Fixed  & 0.0100 &  \textbf{0.0719} & \textbf{0.0200} & 0.0486 \\
        & PINO-SoftAdapt  & 0.0101 & 0.0923  & 0.0216 &  0.0535\\
        & PINO-RelAdapt   & \textbf{0.0095} & 0.0742 & 0.0213 & \textbf{0.0393} \\
    \hline

    \multirow{4}{*}{Spiral-Break} 
        & FNO   & 0.0742 &  0.6321$^{*}$ & 0.1810 & 0.3972$^{*}$ \\
        & PINO-Fixed  & 0.0624 & 0.7444$^{*}$ & 0.1447 & \textbf{0.4140} \\
        & PINO-SoftAdapt  &  0.0654 &  0.6109$^{*}$ &  \textbf{0.1298} &  0.4431  \\
        & PINO-RelAdapt   &  \textbf{0.0577} & \textbf{0.5023} & 0.1490 & 0.4425\\
    \hline
    \hline
    \end{tabular}
    \end{footnotesize}
\label{tab:long_baseline}
\end{table}

\subsection{Training Epoch Variation}
\label{ssec:epoch_tests}
\begin{figure}[h]
  \centering
    \includegraphics[width=\textwidth]{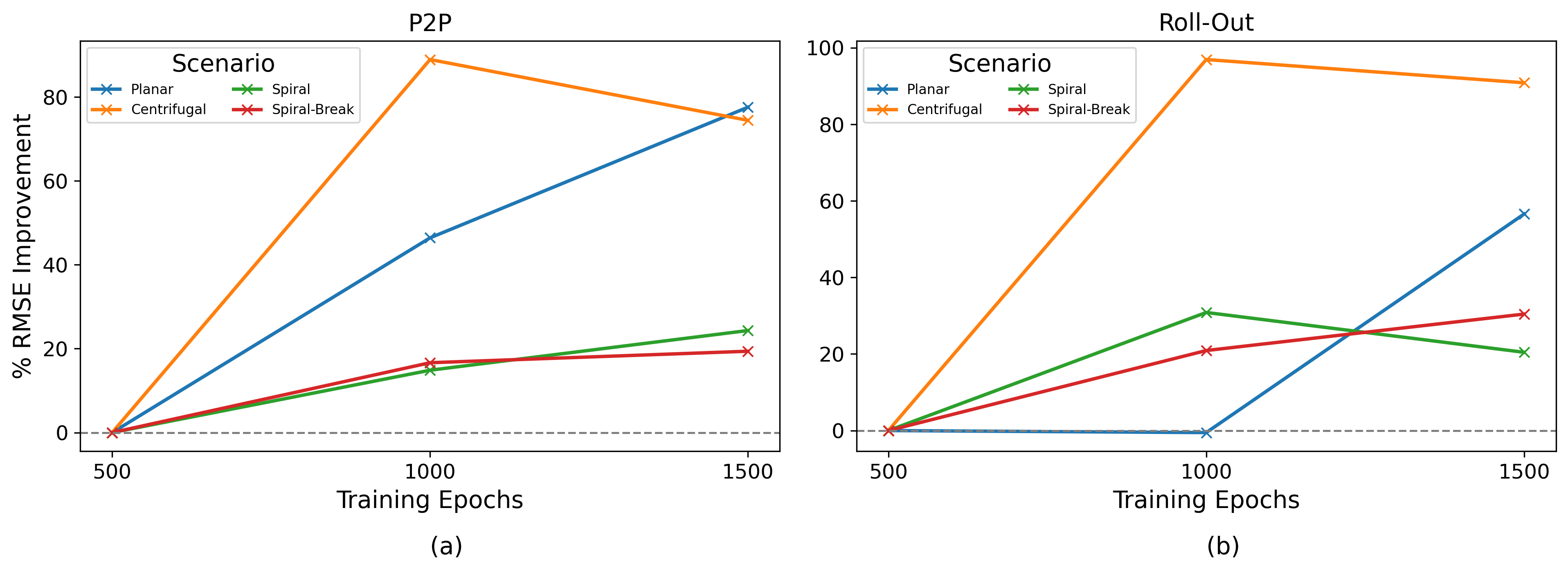}\vspace{0pt}
  \caption{
    Comparison of Model Performance using Multi-Frame Training with training budget, showing the $\%$ RMSE improvement for (a) P2P evaluation, (b) Roll-Out evaluation.
  }
  \label{fig:epoch_comparison}
\end{figure}

We experimented with the number of epochs that were used for the physics-informed training round assess the effect on the model performance. We both reduced the number of epochs that were used from the PINO baseline\footnote{Using the best performing training strategy for each propagation type.} to $500$ and increased it to $1500$. The results can be found in 
Fig~\ref{fig:epoch_comparison}, which demonstrate that in the both the P2P and Roll-Out analysis, reducing the training to $500$ epochs almost universally decreased the performance of the model whilst increasing the number of epochs to $1500$ didn't consistently improve model performance by any significant amount (with the exception of the spiral-break model). These results suggest that increasing the number of epochs can help the model to capture the more complex PDE dynamics of chaotic systems during training whilst for more stable propagation scenarios, increasing the training volume doesn't necessarily improve performance and can occasionally reduce the performance slightly, potentially as a result of over-fitting. In the interest of balancing model accuracy with computational training cost, we chose $1000$ epochs as the baseline training budget.

\subsection{Generalisation Experiments}
\label{ssec:generalisation}
To evaluate the generalisation capabilities of the PINO approach, we compared the performance of models trained on a single propagation scenario with those trained on a mixture of propagation scenarios and model parametrisations. When training these models, the number of epochs was increased to $10000$ in order to allow the model more time to adapt to the wider range of sample types in the dataset.

\subsubsection{Parameter Variation}
\label{sssec:cv_type}
To evaluate the performance of the model on unseen parametrisations of the cell model,  we varied the (isotropic) conductivity of the tissue, which is equivalent to a variation of the diffusion coefficient represented by the parameter $D$ in Eq. \ref{eq:alievpanfilov1}, from the baseline value used in the main body of the text by factors in the range $0.5$ to $1.5$. Here, we considered the spiral model in particular, as the change in the conductivity resulted in significant variation in the movement of the spiral centre and the period of rotation. The training data was generated by varying the conductivity to create $11$ different spiral trajectories, eight of which were selected at random to form the single frame input-output samples for training, using the same temporal splitting for the testing and training (training with the first 80$\%$ of the trajectory and testing on the last 20$\%$) as described in \ref{sec:Experiments}. The remaining three trajectories were held out as evaluation sets to assess the model's ability to adapt to out of distribution parametrisation. 

During training, the parameters for each sample were passed to the model for use in the physics loss calculation; and at inference time the model was assessed using P2P evaluation on the full unseen trajectories. The model's generalisation performance is shown in Table \ref{tab:mixed_cv_spiral}, where we compare the errors for the evaluation trajectories, representing out of distribution (OOD) parametrisation, to those calculated for the test set, representing in distribution (ID) parametrisation.

\begin{table}[h]
\centering
\caption{Performance on spiral propagation under conductivity variation. Evaluation errors are computed over unseen parametrisations (OOD), while test errors correspond to held-out temporal segments of seen trajectories (ID). $\Delta$ and $\Delta\%$ denote the absolute and relative generalization gap (Eval $-$ Test), respectively.}
\label{tab:mixed_cv_spiral}
\begin{footnotesize}
\setlength{\tabcolsep}{5pt}
\renewcommand{\arraystretch}{1.15}
\begin{tabular}{l | c | c | c}
\hline\hline
& \multicolumn{1}{c|}{\textbf{Evaluation (Unseen $D$)}} 
& \multicolumn{1}{c|}{\textbf{Test (Seen $D$)}} 
& \multicolumn{1}{c}{\textbf{$\Delta_{(Eval - Test)}$ } (\textbf{$\Delta\%$})} \\
\hline\hline

Mean RMSE
& 0.00368 
& 0.00280 
& \multicolumn{1}{c}{\textbf{0.00088} (\textbf{31.4\%})} \\


\hline\hline
\end{tabular}
\end{footnotesize}
\end{table}

Although both the evaluation and test RMSE values remain low ($< 4 \times 10^{-3}$), there is a noticeable degradation in performance for the OOD trajectories. This suggests that, despite strong interpolation capabilities over seen conductivity regimes, the model exhibits reduced robustness when extrapolating to unseen parametrisations. One contributing factor may be the random selection of held-out trajectories, which could lead to a test set that is not sufficiently challenging or diverse. Additionally, the observed gap may indicate a degree of over-fitting to the parametrisations seen in training-testing dataset. 

\subsubsection{Zero-Shot Transfer}
\label{sssec:zero_shot_2}
As referenced in Section~\ref{ssec:zero_shot}, we performed additional experiments to evaluate the model's ability to perform predictions on unseen datasets. For these experiments, we trained models across multiple conductivity parametrisations for both the centrifugal and spiral models, this time using single frame samples taken from the full trajectories of each propagation type, with the test set consisting of samples from $2$ of the same held out trajectories from the conductivity variations. At inference time, each model was evaluated on the remaining single unseen trajectory using P2P evaluation for both the propagation type that it was trained on and the unseen propagation type. The results of this evaluation can be found in Table \ref{tab:zero_shot_transfer}\footnote{We note that this evaluation is OOD in terms of conductivity parametrisation for the evaluation that is ID for the propagation type}.

\begin{table}[h]
\centering
\caption{Performance comparison under parameter variation, including zero-shot transfer between propagation types. Error metrics are computed over full trajectories. Diagonal entries correspond to in-domain (ID) evaluation with respect to the propagation type; off-diagonal entries indicate zero-shot transfer between distinct propagation scenarios (OOD). $\Delta$ and $\Delta\%$ denote the absolute and relative generalization gap (OOD $-$ ID), respectively.}
\label{tab:zero_shot_transfer}
\begin{footnotesize}
\setlength{\tabcolsep}{5pt}
\renewcommand{\arraystretch}{1.15}
\begin{tabular}{l | c | c | c }
\hline\hline
\textbf{Trained On} 
& \multicolumn{1}{c|}{\textbf{Centrifugal Eval (RMSE)}} 
& \multicolumn{1}{c|}{\textbf{Spiral Eval (RMSE)}} 
& \multicolumn{1}{c}{\textbf{$\Delta_{(OOD - ID)}$ (\textbf{$\Delta\%$})}} \\
\hline\hline

Centrifugal 
& 0.0027
& 0.0679
& \multicolumn{1}{c}{\textbf{0.0652 ( 2415\%)}}\\


Spiral 
& 0.0134 
& 0.0037
&\multicolumn{1}{c}{\textbf{0.0097 (262\%)}} \\
\hline\hline
\end{tabular}
\end{footnotesize}
\end{table}

Consistent with the previous parametrisation generalisation experiments, whilst the RMSE results for both the ID and OOD evaluation are low ($< 7 \times 10^{-2}$), there is a noticeable reduction in performance for the zero-shot prediction across the propagation type - especially for the centrifugal to spiral model. This result seems reasonable due to the spiral model being a more complex propagation scenario compared to the centrifugal propagation type. 

As noted in Sections~\ref{ssec:zero_shot}, the dominant source of error in the zero-shot setting lies in the amplitude of the predicted voltage rather than in the qualitative structure of the propagation patterns. This indicates that the model tends to successfully capture the underlying dynamics but struggles with accurate scaling for unseen propagation types, showing that there is scope to make improvements to the training procedure (such as enhanced normalisation, loss re-weighting, or amplitude-aware objectives) to improve zero-shot generalisation.

\end{document}